\pdfoutput=1
\documentclass[11pt,a4paper]{article}

\usepackage[hyperref]{acl2021}

\usepackage{header}
\usepackage{times}
\usepackage{latexsym}

\usepackage[T1]{fontenc}
\usepackage[utf8]{inputenc}

\usepackage{microtype}

\aclfinalcopy % Uncomment this line for the final submission
\def\aclpaperid{3407} %  Enter the acl Paper ID here

\title{
Compositional Generalization and Natural Language Variation: \\
Can a Semantic Parsing Approach Handle Both?}
\author{Peter Shaw \quad Ming-Wei Chang \quad Panupong Pasupat \quad Kristina Toutanova \\\\
 Google Research \\
 {\tt \{petershaw,mingweichang,ppasupat,kristout\}@google.com} \\}

\begin{document}

\maketitle
\begin{abstract}

Sequence-to-sequence models excel at handling natural language variation, but have been shown to struggle with out-of-distribution compositional generalization. This has motivated new specialized architectures with stronger compositional biases, but most of these approaches have {\em only} been evaluated on synthetically-generated datasets, which are not representative of natural language variation. 
In this work we ask:
{can we develop a semantic parsing approach that handles both natural language variation and compositional generalization?} To better assess this capability, we propose new train and test splits of non-synthetic datasets. We demonstrate that strong existing approaches do not perform well across a broad set of evaluations. We also propose \qgxt, a hybrid model that combines a high-precision grammar-based approach
with a pre-trained sequence-to-sequence model. It outperforms existing approaches across several compositional generalization challenges on non-synthetic data, while also being competitive with the state-of-the-art on standard evaluations. While still far from solving this problem,
our study highlights the importance of diverse evaluations and the open challenge of handling both compositional generalization and natural language variation in semantic parsing.

\end{abstract}

\section{Introduction}

\begin{figure}[!t]

\begin{center}
\scalebox{.83}{
\begin{tikzpicture}

\begin{scope}[every node/.style={align=center, scale=0.8}]

% Lower-left.
\draw[fill=lgrey,draw=none] (1,1) rectangle (4,2.6);

% upper-left.
\node at (2.5,3.9) {Specialized\\architectures\\with strong\\compositional bias};

% upper-right.
\draw[fill=lyellow,draw=none] (4,2.6) rectangle (7,5.2);
\node at (5.5,3.9) {Under-explored};

% lower-right.
\node at (5.5,1.8) {General-purpose\\pre-trained models\\ (e.g. seq2seq)};

\end{scope}

% title
\node[draw] at (4,5.6) {\textsc{predominant approaches}};

% border and axes
\draw[draw=dgrey,thin] (1,1) rectangle (7,5.2);
\draw[draw=dgrey,thin,dashed] (4,1) -- (4,5.2);
\draw[draw=dgrey,thin,dashed] (1,2.6) -- (7,2.6);

% arrows
\draw[draw=dgrey, very thick, ->,>=latex] (0.85,0.35) -- (0.85,5.3);
\draw[draw=dgrey, very thick, ->,>=latex] (0.85,0.35) -- (7.1,0.35);

% x-axis
\draw[draw=dgrey, thin] (1,0.5) rectangle (7,1);
\node[align=center] at (2.5,0.75) {\small \textsc{synthetic}};
\node[align=center] at (5.5,0.75) {\small \textsc{non-synthetic}};

\node[align=center] at (4,0) {\textsc{natural language variation}};

% y-axis
\node[rotate=90,align=center] at (0.25,2.78) {\textsc{compositional}\\ \textsc{generalization}};

\end{tikzpicture}
}
\end{center}
\caption{
We study whether a semantic parsing approach can handle both out-of-distribution compositional generalization and natural language variation. Existing approaches are commonly evaluated across only one dimension.
}

\label{fig:challenge}
\end{figure}
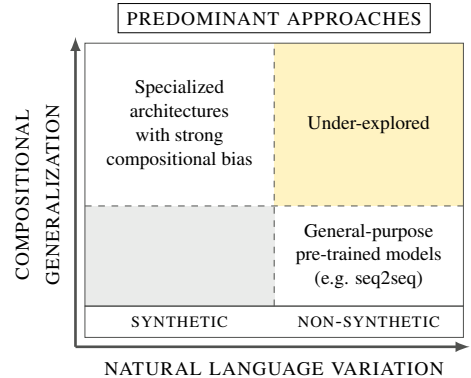

Sequence-to-sequence (seq2seq) models have been widely used in semantic parsing~\cite{dong2016language, jia2016data} and excel at handling the natural language variation\footnote{We use the term \emph{natural language variation} in a broad sense to refer to the many different ways humans can express the same meaning in natural language, including differences in word choice and syntactic constructions.} of human-generated queries. However, evaluations on synthetic\footnote{We make a coarse distinction between \emph{synthetic} datasets, where natural language utterances are generated by a program, and \emph{non-synthetic} datasets, where natural language utterances are collected from humans.}
tasks such as \textsc{SCAN}~\cite{lake2018generalization} have shown that seq2seq models often generalize poorly to out-of-distribution compositional utterances, such as ``jump twice'' when only ``jump'', ``walk'', and ``walk twice'' are seen during training. This ability to generalize to novel combinations of the elements observed during training is referred to as \emph{compositional generalization}.

This has motivated many specialized architectures that improve peformance on \textsc{SCAN}~\cite{li2019compositional,russin2019compositional,gordon2019permutation,lake2019compositional,liu2020compositional,nye2020learning,chen2020compositional}. However, most approaches have only been evaluated on synthetic datasets.
While synthetic datasets enable precise, interpretable evaluation of specific phenomena, they are less representative of the natural language variation that a real-world semantic parsing system must handle.

In this paper, we ask: {\em can we develop a semantic parsing approach that handles both natural language variation and compositional generalization?} Surprisingly, this question is understudied. As visualized in Figure~\ref{fig:challenge}, most prior work evaluates \emph{either} out-of-distribution compositional generalization on synthetic datasets, \emph{or} in-distribution performance on non-synthetic datasets.
Notably, designing approaches that can handle both compositional generalization and the natural language variation of non-synthetic datasets is difficult.\eat{, as different approaches have their own strengths and limitations.} For example, large pre-trained seq2seq models that perform well on in-distribution evaluations do not address most of the compositional generalization challenges proposed in \textsc{SCAN}~\cite{furrer2020compositional}.

\begin{figure}[!t]
\begin{center}
    
\scalebox{.83}{
\begin{tikzpicture}

% title

\node[draw] at (4,5.6) {\textsc{train and test splits}};

% border and axes
\draw[draw=dgrey, thin] (1,1) rectangle (7,5.2);
\draw[draw=dgrey, thin,dashed] (4,1) -- (4,5.2);

% arrows
\draw[draw=dgrey, very thick, ->,>=latex] (0.85,0.35) -- (0.85,5.3);
\draw[draw=dgrey, very thick, ->,>=latex] (0.85,0.35) -- (7.1,0.35);

% x-axis
\draw[draw=dgrey, thin] (1,0.5) rectangle (7,1);
\node[align=center] at (2.5,0.75) {\small \textsc{synthetic}};
\node[align=center] at (5.5,0.75) {\small \textsc{non-synthetic}};

\node[align=center] at (4,0) {\textsc{natural language variation}};

% y-axis
\node[rotate=90,align=center] at (0.25,2.78) {\textsc{compositional}\\ \textsc{generalization}};

% Nodes.

\begin{scope}[every node/.style={rectangle, thick, inner xsep=5pt, inner ysep=4pt, align=center, fill=lblue, draw=dblue, scale=0.8, minimum width=95pt}].

% Synthetic column.
\node at (2.5,4.65) {MCD\\[-0.15cm]~\tiny{~\cite{keysers2019measuring}}};
\node at (2.5,3.76) {Add Primitive\\[-0.15cm]~\tiny{~\cite{lake2018generalization}}};
\node at (2.5,3.0) {Length};

% Non-synthetic column.
\node[fill=lgreen,draw=dgreen, minimum height=24pt] at (5.5,4.62) {TMCD};
\node at (5.5,3.76) {Template\\[-0.15cm]~\tiny{~\cite{finegan2018improving}}};
\node at (5.5,3.0) {Length};
\node at (5.5,1.35) {Random};

% Horizontal dividers.
\draw[draw=dgrey,thin,dashed] (1,2.6) -- (7,2.6);

\end{scope}
\end{tikzpicture}
}
\end{center}

\caption{
We evaluate semantic parsing approaches across a diverse set of evaluations focused on  natural language variation, compositional generalization, or both. We add TMCD splits to complement existing evaluations. Ordering within each cell is arbitrary.
}

\label{fig:evaluations}
\end{figure}
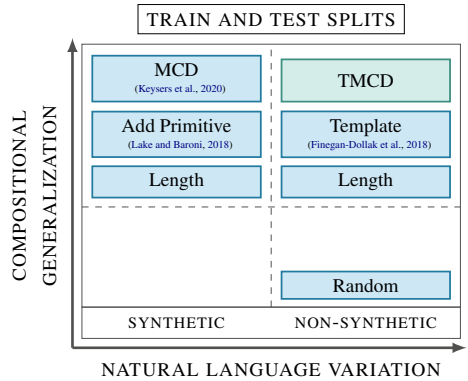

Our research question has two important motivations. First, humans have been shown to be adept compositional learners~\cite{lake2019human}. Several authors have argued that a greater focus on compositional generalization is an important path to more human-like generalization and NLU~\cite{lake2017building,battaglia2018relational}. 
Second, it is practically important to assess performance on non-synthetic data and out-of-distribution examples, as random train and test splits can overestimate real-world performance and miss important error cases~\cite{ribeiro2020beyond}. Therefore, we are interested in approaches that do well not only on controlled synthetic challenges of compositionality or in-distribution natural utterances, but across all of the diverse set of evaluations shown in Figure~\ref{fig:evaluations}.

Our contributions are two-fold. First, on the evaluation front,
we show that performance on \textsc{SCAN} is not well-correlated with performance on non-synthetic tasks. In addition, strong existing approaches do not perform well across all  evaluations in Figure~\ref{fig:evaluations}.
We also propose new Target Maximum Compound Divergence (\tmcd) train and test splits, extending the methodology of~\citet{keysers2019measuring} to create challenging evaluations of compositional generalization for non-synthetic datasets. We show that \tmcd splits complement existing evaluations by focusing on different aspects of the problem.

Second, on the modeling front, we propose \qg, a simple and general grammar-based approach that solves \textsc{SCAN} and also scales to natural utterances, obtaining high precision for non-synthetic data. In addition, we introduce and evaluate  \qgxt, a hybrid model that combines \qg  with T5~\cite{raffel2019exploring}, leading to improvements across several compositional generalization evaluations while also being competitive on the standard splits of \textsc{GeoQuery}~\cite{zelle1996learning} and \textsc{Spider}~\cite{yu2018spider}. Our results indicate that \qgxt is a strong baseline for our challenge of developing approaches that perform well across a diverse  set of evaluations focusing on either natural language variation, compositional generalization, or both. Comparing five approaches across eight evaluations on \textsc{SCAN} and \textsc{GeoQuery}, its average rank is 1, with the rank of the best previous approach (T5) being 2.9; performance is also competitive across several evaluations on \textsc{Spider}.

While still far from affirmatively answering our research question, 
our study highlights the importance of a diverse set of evaluations and the open challenge of handling both compositional generalization and natural language variation.\footnote{Our code and data splits are available at \url{https://github.com/google-research/language/tree/master/language/nqg}.}

\section{Background and Related Work}

In this section, we survey recent work related to 
compositional generalization in semantic parsing.

\paragraph{Evaluations}

To evaluate a model's ability to generalize to novel compositions, previous work has proposed several methods for generating train and test splits, as well as several synthetic datasets.

A widely used synthetic dataset for assessing compositional generalization is \textsc{SCAN}~\cite{lake2018generalization}, which consists
of natural language commands (e.g., ``jump twice'') mapping to action sequences (e.g., ``\nlpx{I\_JUMP I\_JUMP}''). One split for SCAN is the length split, where examples are separated by length such that the test set contains longer examples than the training set. Another is the primitive split, where a given primitive (e.g., ``jump'') is seen
by itself during training, but the test set consists of the primitive recombined with other elements observed during training (e.g., ``jump twice'').
Other synthetic datasets have been developed to evaluate aspects of compositional generalization beyond SCAN, including NACS~\cite{bastings2018jump}, CFQ~\cite{keysers2019measuring}, and COGS~\cite{kim2020cogs}. 

In addition to introducing the CFQ dataset, ~\citet{keysers2019measuring} propose Maximum Compound Divergence (MCD) splits based on the notion of a compound distribution. Their algorithm generates
train and test splits that maximize the divergence of their respective compound distributions while bounding the divergence of their respective atom distributions.
We extend their methodology to create new \tmcd splits for non-synthetic datasets. 

Another method for generating train and test splits is the template\footnote{Also referred to as a \emph{query} split.} split~\cite{finegan2018improving}. Unlike the aforementioned evaluations, template splits have been applied to non-synthetic datasets, primarily for text-to-SQL. In template splits, any parse template (defined as the target SQL query with entities anonymized) appearing in the training set cannot appear in the test set. 
We analyze and discuss template splits in \S~\ref{sec:analysis-tmcd}. 

Finally, ~\citet{herzig2019don} studies biases resulting from methods for efficiently collecting human-labeled data, providing further motivation for out-of-distribution evaluations.

\paragraph{Approaches}

Many specialized architectures have been developed to address the compositional generalization challenges of \textsc{SCAN}. Several of them have recently reached 100\% accuracy across multiple \textsc{SCAN} challenges~\cite{liu2020compositional,nye2020learning,chen2020compositional}. Similarly to the \qgxt approach we propose in \S~\ref{sec:model}, all of these models incorporate discrete structure. However, unlike \qgxt, they have only been evaluated on synthetic parsing tasks.

Recently, \citet{herzig2020span} also begins to address our research question, proposing an approach that not only solves several \textsc{SCAN} challenges but also achieves strong performance on the standard and template splits of the non-synthetic dataset \textsc{GeoQuery}. However, their approach requires some manual task-specific engineering. We compare \qgxt with this approach and other \textsc{SCAN}-inspired architectures. \citet{oren2020improving} and \citet{zheng2020compositional} also explored compositional generalization on non-synthetic datasets by focusing on the template splits proposed by ~\citet{finegan2018improving}, demonstrating improvements over standard seq2seq models.

The effect of large-scale pre-training on compositional generalization ability has also been studied.  \citet{furrer2020compositional} finds that pre-training alone cannot solve several compositional generalization challenges, despite
its effectiveness across NLP tasks such as question answering~\cite{raffel2019exploring}.

While our work focuses on modeling approaches, compositional data augmentation techniques have also been proposed~\cite{jia2016data, andreas2019good}. \qgxt outperforms previously reported results for these methods, but more in-depth analysis is needed.

\section{Target Maximum Compound Divergence (\tmcd) Splits}
\label{sec:tmcd}

The existing evaluations targeting compositional generalization for non-synthetic tasks are template splits and length splits.  Here we propose an additional method 
which expands the set of available evaluations by generating data splits that maximize compound divergence over non-synthetic datasets, termed Target Maximum Compound Divergence (\tmcd) splits.  As we show in \S~\ref{sec:analysis}, it results in a generalization problem with different characteristics that can be much more challenging than template splits, and contributes to the comprehensiveness of evaluation.

In standard MCD splits \cite{keysers2019measuring}, the notion of compounds requires that both source and target are generated by a rule-based procedure, 
and therefore cannot be applied to existing non-synthetic datasets where natural language utterances are collected from humans. For \tmcd, we 
propose a new notion of compounds based only on the target representations. We leverage their known syntactic structure to define atoms and compounds. For instance, example atoms in FunQL are  $\nlp{longest}$ and $\nlp{river}$, and an example compound is $\nlp{longest(river)}$.
Detailed definitions of atoms and compounds for each dataset we study can be found in Appendix~\ref{sec:appendix-tmcd-definitions}. 

Given this definition of compounds, our definition of compound divergence, $\mathcal{D}_C$, is the same as that of~\citet{keysers2019measuring}. 
Specifically,
$$ \mathcal{D}_C
= 1\,-\, C_{0.1}(\mathcal{F}_\textsc{train} \, \Vert \, \mathcal{F}_\textsc{test}),$$
where $\mathcal{F}_\textsc{train}$ and $\mathcal{F}_\textsc{test}$ are the weighted frequency distributions of compounds in the training and test sets, respectively.
The Chernoff coefficient
$C_\alpha(P \Vert Q) = \sum_{k} p_k^\alpha \, q_k^{1-\alpha} $~\cite{chung1989measures} is used with $\alpha = 0.1$.

For \tmcd, we constrain atom divergence by requiring that every atom appear at least once in the training set. An atom constraint is desirable so that the model knows the possible target atoms to generate.
A greedy algorithm similar to the one of~\citet{keysers2019measuring} is used to generate splits that approximately maximize compound divergence. First, we randomly split the dataset. Then, we swap examples until the atom constraint is satisfied. Finally, we sequentially identify example pairs that can be swapped between the train and test sets to increase compound divergence without violating the atom constraint, breaking when a swap can no longer be identified. 

\section{Proposed Approach: \qgxt}
\label{sec:model}

We propose \qgxt, a hybrid semantic parser that combines a grammar-based approach with a seq2seq model. The two components are motivated by prior work focusing on compositional generalization and natural language variation, respectively, and we show in \S~\ref{sec:experiments} that their combination sets a strong baseline for our challenge. 

\begin{figure}[!t]
\begin{center}

\scalebox{0.7}{

\begin{tikzpicture}
\begin{scope}[every node/.style={inner xsep=5pt, inner ysep=5pt}]

\tikzstyle{startstop} = [rectangle, rounded corners,text centered, draw=dred, fill=lred]
\tikzstyle{process} = [rectangle, text centered, draw=dgreen, fill=lgreen]
\tikzstyle{decision} = [diamond, text centered, draw=dblue, fill=lblue, minimum width=3cm, inner sep=-.5ex]

\tikzstyle{arrow} = [thick,->,>=latex]

\node (start) [startstop] {Start};

\node (qg) [process, right=0.5cm of start,align=center] {Run\\\qg};
\node (dec) [decision, right=0.5cm of qg,align=center] {\qg has\\output?};
\node (t5) [process, right=1.25cm of dec,align=center] {Run\\T5};
\node (qgout) [startstop, below=0.75cm of dec] {Return \qg output};
\node (t5out) [startstop, below=0.5cm of t5] {Return T5 output};

\draw [arrow] (start) -- (qg);
\draw [arrow] (qg) -- (dec);
\draw [arrow] (dec) -- node[anchor=south] {no} (t5);
\draw [arrow] (dec) -- node[anchor=east] {yes} (qgout);
\draw [arrow] (t5) -- (t5out);

\end{scope}
\end{tikzpicture}

}

\end{center}

\caption{
Overview of how predictions are generated by \qgxt, a simple yet effective combination of T5~\cite{raffel2019exploring} with a high-precision grammar-based approach, ~\qg.
}
\label{t5_combo}

\end{figure}
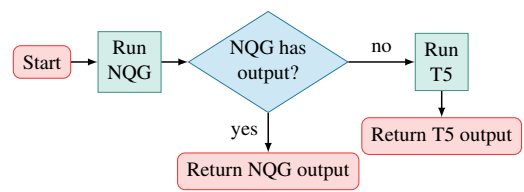

The grammar-based component, \textbf{\qg}, consists of a discriminative \textbf{N}eural parsing model and a flexible \textbf{Q}uasi-synchronous \textbf{G}rammar induction algorithm which can operate over arbitrary pairs of strings. Like other grammar-based approaches, \qg can fail to produce an output for certain inputs. As visualized in Figure~\ref{t5_combo}, in cases where \qg fails to produce an output, we return the output from T5~\cite{raffel2019exploring}, a pre-trained seq2seq model. This simple combination can work well because \qg often has higher precision than T5 for cases where it produces an output, especially in out-of-distribution settings.

We train \qg and T5 separately. Training data for both components consists of pairs of source and target strings, referred to as $\mathbf{x}$ and $\mathbf{y}$, respectively.

\subsection{\qg Component}
\label{sec:nqg-component}

\qg is inspired by more traditional approaches to semantic parsing based on grammar formalisms such as CCG~\cite{zettlemoyer2005learning,zettlemoyer2007online,kwiatkowski2010inducing,kwiatkowski2013scaling} and SCFG~\cite{wong2006learning, wong2007learning,andreas-etal-2013-semantic,li-etal-2015-improving-semantic}.
\qg combines a QCFG induction algorithm with a neural parsing model. Training is a two-stage process.
First, we employ a compression-based grammar induction technique to construct our grammar.  Second, based on the induced grammar, we build the \qg semantic parsing model via a discriminative latent variable model, using a powerful neural encoder to score grammar rule applications anchored in the source string $\mathbf{x}$.

\subsubsection{\qg Grammar Induction}
\label{sec:nqg-grammar-induction}

\paragraph{Grammar Formalism} Synchronous context-free grammars (SCFGs) \emph{synchronously} generate strings in both a source and target language.
Compared to related work based on SCFGs for machine translation~\cite{chiang2007hierarchical} and semantic parsing,
\qg uses a slightly more general grammar formalism that allows repetition of a non-terminal with the same index on the target side. Therefore,
we adopt the terminology of quasi-synchronous context-free grammars~\cite{smith2006quasi}, or QCFGs, to refer to our induced grammar $\mathcal{G}$.\footnote{See Appendix~\ref{sec:appendix-qcfg-background} for additional background on QCFGs.} Our grammar $\mathcal{G}$ contains a single non-terminal symbol, $NT$. We restrict source rules to ones containing at most 2 non-terminal symbols, and do not allow unary productions as source rules. This enables efficient parsing using an algorithm similar to CKY~\cite{cocke1969programming,kasami1965,younger1967recognition} that does not require binarization of the grammar.

\paragraph{Induction Procedure}

To induce $\mathcal{G}$ from the training data, we propose a QCFG induction algorithm that does not rely on task-specific heuristics or pre-computed word alignments. Notably, our approach makes no explicit assumptions about the source or target languages, beyond those implicit in the QCFG formalism. Table~\ref{rule_examples} shows examples of induced rules.

\begin{table}[t!]

\begin{center}
\scalebox{0.79}{
\begin{tabular}{l}

\toprule
\bf $\textsc{SCAN}$ \\[0.1cm]
$ NT \rightarrow \langle \textrm{turn right}, \textrm{I\_TURN\_RIGHT} \rangle $ \\ 
$ NT \rightarrow \langle NT_{\boxnum{1}}~\textrm{after}~NT_{\boxnum{2}}, NT_{\boxnum{2}}~NT_{\boxnum{1}} \rangle $ \\
$ NT \rightarrow \langle NT_{\boxnum{1}}~\textrm{thrice}, NT_{\boxnum{1}}~NT_{\boxnum{1}}~NT_{\boxnum{1}} \rangle $ \\
\midrule
\bf $\textsc{GeoQuery}$ \\[0.1cm]
$ NT \rightarrow \langle \textrm{names of}~NT_{\boxnum{1}}, NT_{\boxnum{1}} \rangle $ \\  
$ NT \rightarrow \langle \textrm{towns}, \textrm{cities} \rangle $ \\ 
$ NT \rightarrow \langle NT_{\boxnum{1}}~\textrm{have}~NT_{\boxnum{2}}~\textrm{running through them}, $ \\
$ ~~~~~~~~~~~~~~\textrm{intersection (}~NT_{\boxnum{1}}~\textrm{, traverse\_1 (}~NT_{\boxnum{2}}~\textrm{) )} \rangle $ \\
\midrule
\bf $\textsc{Spider-SSP}$ \\[0.1cm]
$ NT \rightarrow \langle \textrm{reviewer}, \textrm{reviewer} \rangle $ \\ 
$ NT \rightarrow \langle \textrm{%movie\_1: 
what is the id of the}~NT_{\boxnum{1}}~\textrm{named}~NT_{\boxnum{2}}~\textrm{?}, $ \\
$ ~~~~~~~~~~~~~~\textrm{select rid from}~NT_{\boxnum{1}}~\textrm{where name = "}~NT_{\boxnum{2}}~\textrm{"} \rangle $ \\
\bottomrule

\end{tabular}
}
\end{center}
\caption{Examples of induced QCFG rules. The subscript $1$ in
$NT_{\boxnum{1}}$ indicates the correspondence between source and target non-terminals.}
\label{rule_examples}
\end{table}

Our grammar induction algorithm is guided by the principle of Occam's razor, which leads us to seek the smallest, simplest grammar that explains the data well. We follow the Minimum Description Length (MDL) principle~\cite{rissanen1978modeling, grunwald2004tutorial} as a way to formalize this intuition. 
Specifically, we use standard two-part codes to compute description length, where we are interested in an encoding of targets $\mathbf{y}$ given the inputs $\mathbf{x}$, across a dataset $\mathcal{D}$ consisting of these pairs. A two-part code encodes the model and the targets encoded using the model; the two parts measure the simplicity of the model and the extent to which it can explain the data, respectively.

For grammar induction, our model is simply our grammar, $\mathcal{G}$. The codelength can therefore be expressed as $H(\mathcal{G}) - \sum_{\mathbf{x},\mathbf{y} \in \mathcal{D}} \log_2 
P_{\mathcal{G}}( \mathbf{y} | \mathbf{x} )$ where $H(\mathcal{G})$ corresponds to the codelength of some encoding of $\mathcal{G}$. We approximate $H(\mathcal{G})$ by counting terminal ($C_T$) and non-terminal ($C_{N}$) symbols in the grammar's rules, $\mathcal{R}$. For $P_{\mathcal{G}}$, we assume a uniform distribution over the set of possible derivations.\footnote{This can be viewed as a conservative choice, as in practice we expect our neural parser to learn a better model for $P(\mathbf{y}|\mathbf{x})$ than a naive uniform distribution over derivations.} As the only mutable aspect of the grammar during induction is the set of rules $\mathcal{R}$, we abuse notation slightly and write our approximate codelength objective as a function of $\mathcal{R}$ only:

\begin{equation*}
\begin{gathered}
L(\mathcal{R}) = l_{N} C_{N}(\mathcal{R}) + l_T C_T(\mathcal{R}) - \\
\sum \limits_{(\mathbf{x},\mathbf{y}) \in \mathcal{D}} \log_2 \frac{|Z^\mathcal{G}_{\mathbf{x},\mathbf{y}}|}{|Z^\mathcal{G}_{\mathbf{x},*}|},
\end{gathered}
\end{equation*}
where $Z^\mathcal{G}_{\mathbf{x},\mathbf{y}}$
is the set of all derivations in $\mathcal{G}$ that yield the pair of strings $\mathbf{x}$ and $\mathbf{y}$, while $Z^\mathcal{G}_{\mathbf{x},*} \supset Z^\mathcal{G}_{\mathbf{x},\mathbf{y}}$ is the set of derivations that yield source string $\mathbf{x}$ and any target string. The constants $l_N$ and $l_T$ can be interpreted as the average bitlength for encoding non-terminal and terminal symbols, respectively. In practice, these are treated as hyperparameters.

We use a greedy search algorithm to find a grammar that approximately minimizes this codelength objective. We initialize $\mathcal{G}$ by creating a rule
$NT \rightarrow \langle \mathbf{x}, \mathbf{y} \rangle$
for every training example $(\mathbf{x}, \mathbf{y})$.
By construction, the initial grammar perfectly fits the training data, but is also very large.
Our algorithm iteratively identifies a rule that can be added to $\mathcal{G}$ that decreases our codelength objective by enabling $\geq 1$ rule(s) to be removed, under the invariant constraint that $\mathcal{G}$ can still derive all training examples. The search completes when no rule that decreases the objective can be identified. In practice, we use several approximations to efficiently select a rule at each iteration. Additional details regarding the grammar induction algorithm are described in Appendix~\ref{sec:qcfq_induction}.

\subsubsection{\qg Semantic Parsing Model} 

Based on the induced grammar $\mathcal{G}$, we train a discriminative latent variable parsing model, using a method similar to that of  \citet{blunsom-etal-2008-discriminative}. We define $p(\mathbf{y} \mid \mathbf{x})$ as:
$$ p(\mathbf{y} \mid \mathbf{x}) = \sum_{z \in Z^\mathcal{G}_{\mathbf{x},\mathbf{y}}} p(\mathbf{z} \mid \mathbf{x}), $$
where $Z^\mathcal{G}_{\mathbf{x},\mathbf{y}}$ is the set of derivations of $\mathbf{x}$ and $\mathbf{y}$ in $\mathcal{G}$. We define ${p(\mathbf{z} \mid \mathbf{x})}$ as:
$$ p(\mathbf{z} \mid \mathbf{x}) = \frac{ \exp(s(\mathbf{z}, \mathbf{x})) }{ \sum\limits_{z^\prime \in Z^\mathcal{G}_{\mathbf{x},*}} \exp(s(\mathbf{z^\prime}, \mathbf{x})) }, $$
where $s(\mathbf{z}, \mathbf{x})$ is a derivation score and the denominator is a global partition function. Similarly to the Neural CRF model of \citet{durrett2015neural}, the scores decompose over anchored rules. Unlike \citet{durrett2015neural}, we compute these scores based on contextualized representations from a BERT~\cite{devlin2018bert} encoder. Additional details regarding the model architecture can be found in Appendix~\ref{sec:appendix-parsing-model}.

At training time, we use a Maximum Marginal Likelihood (MML) objective. We preprocess each example to produce parse forest representations for both $Z^\mathcal{G}_{\mathbf{x},\mathbf{y}}$ and $Z^\mathcal{G}_{\mathbf{x},*}$, which correspond to the numerator and denominator of our MML objective, respectively. By using dynamic programming to efficiently sum derivation scores inside the training loop, we can efficiently compute the exact MML objective without requiring approximations such as beam search. 

At inference time, we select the highest scoring derivation using an algorithm similar to CKY that considers anchored rule scores generated by the neural parsing model. We output the corresponding target if it can be derived by a CFG defining valid target constructions for the given task.

\subsubsection{\qg Discussion} 
We note that \qg is closely related to work that uses synchronous grammars for hierarchical statistical machine translation, such as Hiero~\cite{chiang2007hierarchical}. Unlike Hiero, \qg does not rely on an additional word alignment component.
Moreover, Hiero simply uses relative frequency to learn rule weights. 
Additionally, in contract with traditional SCFG models for machine translation applied to semantic parsing ~\cite{wong2006learning,andreas-etal-2013-semantic}, our neural model conditions on global context from the source $\mathbf{x}$ via contextual word embeddings, and our grammar's rules do not need to carry source context to aid disambiguation.

\subsection{T5 Component}

T5~\cite{raffel2019exploring} is a pre-trained sequence-to-sequence Transformer model~\cite{vaswani2017attention}. We fine-tune T5 for each task.
\section{Experiments}
\label{sec:experiments}

\begin{table*}[!ht]
\begin{center}

\scalebox{0.82}{
\begin{tabular}{lccccccccc}
\toprule
 & \multicolumn{4}{c}{{\bf{\textsc{SCAN}}}} &  \multicolumn{4}{c}{{\bf\textsc{GeoQuery}}} & \multirow{2}{0.9cm}{Avg. Rank} \\ 
 \cmidrule(lr){2-5} \cmidrule(lr){6-9}

System & Jump & Turn Left & Len. & MCD & Standard & Template &  Len. & \tmcd \\
\midrule
LANE ~\cite{liu2020compositional} & 
\cellcolor{cellshade} \bf 100 & 
 --- & 
\cellcolor{cellshade} \bf 100 & 
\cellcolor{cellshade} \bf 100 & --- & --- & --- & --- & --- \\
NSSM ~\cite{chen2020compositional} & 
\cellcolor{cellshade} \bf 100 & 
 --- & 
\cellcolor{cellshade} \bf 100 
& --- & --- & --- & --- & --- & --- \\
 Syntactic Attn. \cite{russin2019compositional} &
\cellcolor{cellshade}  91.0 & 
\cellcolor{cellshade}  \bf 99.9 & 
\cellcolor{cellshade}  15.2 & 2.9 &
77.5 &
70.6 &
23.6 &
\phantom{0}0.0 & 3.9 \\
 CGPS \cite{li2019compositional} & 
\cellcolor{cellshade}  98.8 & 
\cellcolor{cellshade}  \bf 99.7 & 
\cellcolor{cellshade}  20.3 & 
\cellcolor{cellshade}  2.0 & 
62.1 & 
32.8 &
\phantom{0}9.3 & 
32.3 & 4.4 \\
GECA \cite{andreas2019good} &
\cellcolor{cellshade}  87.0 & 
 --- & 
 --- &
 --- &
\cellcolor{cellshade} 78.0$^{\dagger}$ &
  --- &
  --- &
  --- & --- \\
SBSP \cite{herzig2020span} &
\cellcolor{cellshade} \bf  100 & 
\cellcolor{cellshade} \bf 100 & 
\cellcolor{cellshade} \bf 100 &
\cellcolor{cellshade} \bf 100 &
\cellcolor{cellshade} 86.1$^{\dagger}$ &
 --- &
 --- &
 --- & --- \\
SBSP ~$-$\em{lexicon} &
\cellcolor{cellshade} \bf 100 & 
\cellcolor{cellshade} \bf 100 & 
\cellcolor{cellshade} \bf 100 &
\cellcolor{cellshade} \bf 100 &
\cellcolor{cellshade} 78.9$^{\dagger}$ &
 --- &
 --- &
 --- & --- \\
\midrule
T5-Base \cite{raffel2019exploring} & 
\cellcolor{cellshade} \bf 99.5 & 
\cellcolor{cellshade} 62.0 & 
\cellcolor{cellshade} 14.4 & 
\cellcolor{cellshade} 15.4 & 
\bf 92.9 & 
87.0 &
39.1 & 
54.3 & 2.9 \\
T5-3B \cite{raffel2019exploring}& 
\cellcolor{cellshade} \bf 99.0 & 
\cellcolor{cellshade} 65.1 & 
\cellcolor{cellshade} \phantom{0}3.3 & 
\cellcolor{cellshade} 11.6 & 
\bf 93.2 & 
83.1 &
36.8 & 
51.6 & --- \\
\midrule
\qgxt-Base & 
\bf 100 & 
\bf 100 & 
\bf 100 & 
\bf 100 & 
\bf 92.9 & 
\bf 88.8 &
\bf 52.2 & 
\bf 56.6 & \bf 1.0 \\
\qgxt-3B & 
\bf 100 & 
\bf 100 & 
\bf 100 & 
\bf 100 & 
\bf 93.7 & 
85.0 &
\bf 51.4 & 
54.1 & --- \\
\qg & 
\bf 100 & 
\bf 100 & 
\bf 100 & 
\bf 100 & 
76.8 &
61.9 &
37.4 & 
41.1 & 2.3 \\
\bottomrule
\end{tabular}
}
\caption{
{\bf Main Results.} Existing approaches do not excel on a diverse set of evaluations across synthetic and non-synthetic tasks, but
\qgxt obtains significant improvements. For comparison, we report the average rank among 5 approaches across all 8 evaluations. Gray cells are previously reported results. $^{\dagger}$ indicates differences in \textsc{GeoQuery} settings (see discussion in \S~\ref{sec:syn_vs_nonsync}). Boldfaced results are within 1.0 points of the best result. 
}
\label{tab:scan_geo}
\end{center}
\end{table*}

We evaluate existing approaches and the newly proposed \qgxt across a diverse set of evaluations to assess  compositional generalization and handling of natural language variation. We aim to understand how the approaches compare to each other for each type of evaluation and in aggregate, and how the performance of a single approach may vary across different evaluation types.

\subsection{Experiments on \textsc{SCAN} and \textsc{GeoQuery}}
\label{sec:syn_vs_nonsync}

For our main experiments, we focus on evaluation across multiple splits of two datasets with compositional queries: \textsc{SCAN}~\cite{lake2018generalization} and \textsc{GeoQuery}~\cite{zelle1996learning,tang2001using}. The two datasets have been widely used to study compositional generalization and robustness to natural language variation, respectively.  
Both datasets are closed-domain and have outputs with straightforward syntax, enabling us to make clear comparisons between synthetic vs.\ non-synthetic setups.

\paragraph{Approaches}

For \qgxt, to assess the effect of model size, we compare two  sizes of the underlying T5 model: Base (220 million parameters) and 3B (3 billion parameters). To evaluate  \qg individually, we treat any example where no output is provided as incorrect when computing accuracy.

We select strong approaches from prior work that have performed well in at least one setting. We group them into two families of  approaches described in Figure~\ref{fig:challenge}.
First, for general-purpose models that have shown strong ability to  handle natural language variation, we consider T5, a  pre-trained seq2seq model, in both Base and 3B sizes.

Second, for specialized methods with strong compositional biases, we consider approaches that have been developed for \textsc{SCAN}. Some previous approaches for SCAN require task-specific information such as the mapping of atoms \cite{lake2019compositional,gordon2019permutation} or a grammar mimicking the training data \cite{nye2020learning},
and as such are difficult to adapt to non-synthetic datasets.
Among the approaches that do not need task-specific resources, we evaluate two models with publicly available code: Syntactic Attention~\cite{russin2019compositional} and CGPS~\cite{li2019compositional}. We report  results on SCAN from the original papers as well as new results on our proposed data splits.

\paragraph{Datasets} For the \textsc{SCAN} dataset, we evaluate using the length split and two primitive splits, \emph{jump} and \emph{turn left}, included in the original dataset~\cite{lake2018generalization}. We also evaluate using the \textsc{SCAN} MCD splits from \citet{keysers2019measuring}. 
 
\textsc{GeoQuery}~\cite{zelle1996learning} contains natural language questions about US geography. Similarly to prior work~\cite{dong2016language,dong2018coarse}, we 
replace entity mentions with placeholders. We use a variant of Functional Query Language (FunQL) as the target representation~\cite{kate2005learning}.
In addition to the standard split of ~\citet{zettlemoyer2005learning}, we generate multiple splits focusing on compositional generalization:  a new split based on query length and a TMCD split, each consisting of 440 train and 440 test examples. We also generate a new template split consisting of 441 train and 439 test examples.\footnote{We generate a new template split rather than use the \textsc{GeoQuery} template split of \citet{finegan2018improving} to avoid overlapping templates between the train and test sets when mapping from SQL to FunQL.}

We report exact-match accuracy for both datasets.\footnote{For \textsc{GeoQuery} we report the mean of 3 runs for \qg, with standard deviations reported in Appendix ~\ref{sec:appendix-geoquery-variance}} Hyperparameters and pre-processing details can be found in Appendix~\ref{sec:appendix-exp-details}.

\paragraph{Results} 
The results are presented in Table~\ref{tab:scan_geo}.
The results for T5 on SCAN are from~\newcite{furrer2020compositional}.
Additionally, we include results for GECA\footnote{GECA reports \textsc{GeoQuery} results on a setting with Prolog logical forms and without anonymization of entities. Note that the performance of GECA depends on both the quality of the generated data and the underlying parser~\cite{jia2016data}, which can complicate the analysis.}~\cite{andreas2019good},
a data augmentation method,
as well as LANE~\cite{liu2020compositional} and NSSM~\cite{chen2020compositional}\footnote{These SCAN-motivated approaches both include aspects of discrete search and curriculum learning, and have not been demonstrated to scale effectively to non-synthetic parsing tasks. Moreover, the code is either not yet released (NSSM) or specialized to SCAN (LANE).}.
We also compare with SpanBasedSP\footnote{SpanBasedSP preprocesses \textsc{SCAN} to add program-level supervision. For \textsc{GeoQuery}, they similarly use FunQL, but uses slightly different data preprocessing and report denotation accuracy. We computed NQG-T5's denotation accuracy to be 2.1 points \emph{higher} than exact-match accuracy on the standard split of GeoQuery.}~\cite{herzig2020span}.

From the results, we first note that the relative performance of approaches on compositional splits of \textsc{SCAN} is not very predictive of their relative performance on compositional splits of \textsc{GeoQuery}. For example, GGPS is better than T5 on the length split of \textsc{SCAN} but is significantly worse than T5 on the length split of \textsc{GeoQuery}. Similarly, the ranking of most methods is different on the (T)MCD splits of the two datasets.  
Second, the proposed \qgxt approach combines the strengths of T5 and  \qg to achieve superior results across all evaluations. It improves over T5 on compositional generalization for both synthetic and non-synthetic data while maintaining T5's performance on handling in-distribution natural language variation, leading to an average rank of 1.0 compared to 2.9 for T5. (To the best of our knowledge, both T5 and \qgxt achieve new state-of-the-art accuracy on the standard split of \textsc{GeoQuery}.)

Finally, we note that there is substantial room for improvement on handling both compositional generalization and natural language variation.

\subsection{Experiments on \textsc{Spider}}
\label{sec:complicated_datset}

We now compare the approaches on \textsc{Spider}~\cite{yu2018spider}, a non-synthetic text-to-SQL dataset that includes the further challenges of schema linking and modeling complex SQL syntax.

\textsc{Spider} contains 10,181 questions and 5,693 unique SQL queries across 138 domains. The primary evaluation is in the cross-database setting,
where models are evaluated on examples for databases not seen during training. The primary challenge in this setting is generalization to new database schemas,
which is not our focus. Therefore, we use a setting where the databases are shared between train and test examples.\footnote{This is similar to the ``example split'' discussed in ~\citet{yu2018spider}. However, we only consider examples in the original training set for databases with more than 50 examples to ensure sufficient coverage over table and column names in the training data. This includes 51 databases.} We generate 3 new  splits consisting of 3,282 train and 1,094 test examples each: a random split, a split based on source length, and a TMCD split. We also generate a template split by anonymizing integers and quoted strings, consisting of 3,280 train and 1,096 test examples. We adopt the terminology of~\citet{suhr2020exploring} and use \textsc{Spider-SSP} to refer to these same-database splits, and use \textsc{Spider-XSP} to refer to the standard cross-database setting.

We prepend the name of the target database to the source sequence. For T5, we also serialize the database schema as a string and append it to the source sequence similarly to~\citet{suhr2020exploring}. We report exact set match without values, the standard Spider evaluation metric~\cite{yu2018spider}.

\paragraph{Results} Table~\ref{tab:spider-ssp} shows the results of T5 and \qgxt on different splits of \textsc{Spider-SSP}. We also show T5-Base performance without the schema string appended.
The text-to-SQL mapping is not well modeled by \qg.
Nevertheless, the performance of \qgxt is competitive with T5, indicating a strength of the hybrid approach.

Table~\ref{tab:spider-xsp} shows the results on \textsc{Spider-XSP}, which focuses on handling unseen schema rather than compositional generalization. To our surprise, T5-3B proves to be competitive with the state-of-the-art~\cite{choi2020ryansql}
for approaches without access to database contents beyond the table and column names.
As \qgxt simply uses T5's output when the induced grammar lacks coverage, it too is competitive.

\begin{table}[!t]
\begin{center}

\scalebox{0.78}{
\begin{tabular}{lcccc}
\toprule
 & \multicolumn{4}{c}{{\bf\textsc{Spider-SSP}}}\\
 \cmidrule(lr){2-5}
System & Rand. & Templ. & Len. & \tmcd \\
\midrule
T5-Base ~$-$\em{schema} & 
76.5 & 45.3 & 42.5 & 42.3 \\
T5-Base & 
82.0 & 59.3 & 49.0 & 60.9 \\
T5-3B & 
85.6 & 64.8 & 56.7 & 69.6 \\
\midrule
\qgxt-Base & 
81.8 & 59.2 & 49.0 & 60.8 \\
\qgxt-3B & 
85.4 & 64.7 & 56.7 & 69.5 \\
\qg &
\phantom{0}1.3 & \phantom{0}0.5 & \phantom{0}0.0 & \phantom{0}0.5 \\ 
\bottomrule
\end{tabular}
}

\caption{Results on Spider-SSP. While the text-to-SQL task is not modeled well by the \qg grammar due to SQL's complex syntax, \qgxt still performs well by relying on T5.}
\label{tab:spider-ssp}

\end{center}
\end{table}

\begin{table}[!t]
\begin{center}

\scalebox{0.78}{
\begin{tabular}{lc}
\toprule
& {\bf\textsc{Spider-XSP}} \\
System & Dev \\
\midrule
RYANSQL v2~\cite{choi2020ryansql} & 70.6 \\
\midrule
T5-Base &  57.1 \\
T5-3B & 70.0 \\
\midrule
\qgxt-Base & 
57.1 \\
\qgxt-3B & 
70.0 \\
\qg &
\phantom{0}0.0 \\ 
\bottomrule
\end{tabular}
}

\caption{Although Spider-XSP is not our focus, T5 and \qgxt are competitive with the state-of-the-art.}
\label{tab:spider-xsp}
\end{center}
\end{table}

\section{Analysis}
\label{sec:analysis}

\begin{table*}[!t]
\begin{center}

\scalebox{0.78}{
\begin{tabular}{lcccccccccccc}
\toprule
 & \multicolumn{4}{c}{{\bf \textsc{SCAN}}} & \multicolumn{4}{c}{{\bf \textsc{GeoQuery}}} &  \multicolumn{4}{c}{{\bf \textsc{Spider-SSP}}} \\ 
 \cmidrule(lr){2-5} \cmidrule(lr){6-9} \cmidrule(lr){10-13}
Metric & Jump & Turn L. & Len. & MCD & Stand. & Templ. & Len. & \tmcd & Rand. & Templ. & Len. & \tmcd \\
\midrule

\qg Coverage & 
100 &
100 &
100 &
100 &
80.2 & 
64.5 &
43.3 &
43.7 &
\phantom{0}1.5 & 
\phantom{0}0.5 &
\phantom{0}0.0 &
\phantom{0}0.6 \\

\qg Precision &
100 &
100 &
100 &
100 &
95.7 & 
95.8 &
86.4 &
94.1 &
87.5 & 
83.3 &
--- &
85.7
\\

\bottomrule
\end{tabular}
}

\caption{\qg coverage and precision.
\qgxt outperforms T5 when \qg has higher precision than T5 over the subset of examples it covers.}
\label{tab:qg_precision}

\end{center}
\end{table*}

\eat{
\qgxt outperforms T5 on splits where \qg has higher precision than T5 for the subset of examples where \qg produces an output. Table \ref{tab:qg_precision} shows the precision and coverage of \qg, which shows how many outputs are generated by \qg compared to T5 for each split. We analyze the challenge of \tmcd splits in particular, and then analyze each component of \qgxt individually.
}

\subsection{Comparison of Data Splits}
\label{sec:analysis-tmcd}

\begin{table}[ht]
\begin{center}

\scalebox{0.78}{
\begin{tabular}{lcccc}
\toprule
Dataset & Split & $\%_{ZA}$ & $\mathcal{D}_C$ & T5-Base \\
\midrule
\textsc{GeoQuery} & Standard & 0.3 & 0.03 & 92.9 \\
\midrule
\textsc{GeoQuery} & Random & 1.4 & 0.03 & 91.1 \\
\textsc{GeoQuery} & Template & 0.9 & 0.07 & 87.0 \\
\textsc{GeoQuery} & Length & 4.3 & 0.17 & 39.1 \\
\textsc{GeoQuery} & TMCD & 0 & 0.19 & 54.3 \\
\midrule
\textsc{Spider-SSP} & Random & 6.2 & 0.03 & 82.0 \\
\textsc{Spider-SSP} & Template & 30.3 & 0.08 & 59.2 \\
\textsc{Spider-SSP} & Length & 27.4 & 0.08 & 49.0 \\
\textsc{Spider-SSP} & \tmcd & 0 & 0.18 & 60.9 \\
\bottomrule
\end{tabular}
}

\caption{Percentage of test examples with atoms not included in the training set ($\%_{ZA}$), compound divergence ($\mathcal{D}_C$), and T5-Base accuracy for various dataset splits.}
\label{tab:compound_divergence}

\end{center}
\end{table}

Table~\ref{tab:compound_divergence} compares the compound divergence, the number of test examples with unseen atoms, and the accuracy of T5-Base across various splits.
For \textsc{GeoQuery}, the \tmcd split is significantly more challenging than the template split. However, for \textsc{Spider}, the template and \tmcd splits are similarly challenging. Notably, template splits do not have an explicit atom constraint. We find that for the \textsc{Spider} template split, T5-Base accuracy is 53.9\% for the 30.3\% of test set examples that contain an atom not seen during training, and 61.6\% on the remainder, indicating that generalization to unseen atoms can contribute to the difficulty of template splits.\footnote{Future work could explore different choices for constructing template and \tmcd splits, such as alternative compound definitions and atom constraints.} Length splits are also very challenging, but they lead to a more predictable error pattern for seq2seq models, as discussed next.

\subsection{T5 Analysis}

We analyze \qgxt's components, starting with T5. On length splits, there is a consistent pattern to the errors. T5's outputs on the test set are not significantly longer than the maximum length observed during training, leading to poor performance. This phenomenon was explored by \citet{newman2020eos}.

Diagnosing the large generalization gap on the (T)MCD splits is more challenging, but we noticed several error patterns.
For T5-Base on the \textsc{GeoQuery} \tmcd split, in 52 of the 201 incorrect predictions (26\%), the first incorrectly predicted symbol occurs when the gold symbol has 0 probability under a trigram language model fit to the training data. This suggests that the decoder's implicit target language model might have over-fitted to the distribution of target sequences in the training data, hampering its ability to generate novel compositions. Non-exclusively with these errors, 53\% of the incorrect predictions occur when the gold target contains an atom that is seen in only 1 example during training, suggesting that T5 struggles with single-shot learning of new atoms.
In other cases, the errors appear to reflect over-fitting to spurious correlations between inputs and outputs. Some error examples are shown in Appendix~\ref{sec:appendix-geoquery-examples}.

\subsection{\qg Analysis}

To analyze \qg, we compute its coverage (fraction of examples where \qg produces an output) and precision (fraction of examples with a correct output among ones where an output is produced) on different data splits.
The results in Table~\ref{tab:qg_precision} show that \qg has high precision but struggles at coverage on some data splits.

There is a significant difference in the effectiveness of the grammar induction procedure among the three datasets.
Induction is particularly unsuccessful for ~\textsc{Spider},
as SQL has complicated syntax and often requires complex coordination across discontinuous clauses. 
Most of the induced rules are limited to simply replacing table and column names or value literals with non-terminals, such as the rule shown in Table~\ref{rule_examples}, rather than representing nested sub-structures. The degree of span-to-span correspondence between natural language and SQL is seemingly lower than for other formalisms such as FunQL, which limits the effectiveness of grammar induction. Intermediate representations for SQL such as SemQL~\cite{guo2019towards} may help increase the correspondence between source and target syntax.

For both ~\textsc{GeoQuery} and ~\textsc{Spider}, \qg is limited by the expressiveness of QCFGs and the simple greedy search procedure used for grammar induction, which can lead to sub-optimal approximations of the induction objective. Notably, QCFGs cannot directly represent relations between source strings, such as semantic similarity, or relations between target strings, such as logical equivalence (e.g. \nlp{intersect(a,b)} $\Leftrightarrow$ \nlp{intersect(b,a)}), that could enable greater generalization. However, such extensions pose additional scalability challenges, requiring new research in more flexible approaches for both learning and inference.
\section{Conclusions}

Our experiments and analysis demonstrate that NQG and T5 offer different strengths. NQG generally has higher precision for out-of-distribution examples, but is limited by the syntactic constraints of the grammar formalism and by requiring exact lexical overlap with induced rules in order to provide a derivation at inference time. T5's coverage is not limited by such constraints, but precision can be significantly lower for out-of-distribution examples.
With NQG-T5, we offer a simple combination of these strengths. While accuracy is still limited for out-of-distribution examples where NQG lacks coverage, we believe it sets a strong and simple baseline for future work.

More broadly, our work highlights that evaluating on a diverse set of benchmarks is important, and that handling both out-of-distribution compositional generalization and natural language variation remains an open challenge for semantic parsing.  

\section*{Acknowledgements}

We thank Kenton Lee, William Cohen, Jeremy Cole, and Luheng He for helpful discussions. Thanks also to Emily Pitler, Jonathan Herzig, and the anonymous reviewers for their comments and suggestions.

\section*{Ethical Considerations}

This paper proposed to expand the set of benchmarks used to evaluate compositional generalization in semantic parsing. 
While we hope that ensuring semantic parsing approaches perform well across a diverse set of evaluations, including ones that test out-of-distribution compositional generalization, would lead to systems that generalize better to languages not well represented in small training sets, we have only evaluated our methods on semantic parsing datasets in English.

Our \qgxt method uses a pre-trained T5 model, which is computationally expensive in fine-tuning and inference, especially for larger models (see Appendix B.1 for details on running time and compute architecture). Our method does not require pre-training of large models, as it uses pre-existing model releases. \qgxt-base outperforms or is comparable in accuracy to T5-3B on the non-SQL datasets, leading to relative savings of computational resources.

\bibliography{main}
\bibliographystyle{acl_natbib}

\newpage

\appendix 

\clearpage

\begin{center}
{\bf \large{Appendix}}
\end{center}

We organize the appendix into two sections:
\begin{itemize}
   \item Additional details for \qg in Appendix~\ref{sec:appendix-nqg-details}.
   
   \item Additional experimental details and analysis in Appendix~\ref{sec:appendix-exp-details}.
\end{itemize}

\section{\qg Details}
\label{sec:appendix-nqg-details}

In this section we describe the \qg grammar induction algorithm and parsing model in detail, starting with relevant background and notation for QCFGs.

\subsection{Background: SCFGs and QCFGs}
\label{sec:appendix-qcfg-background}

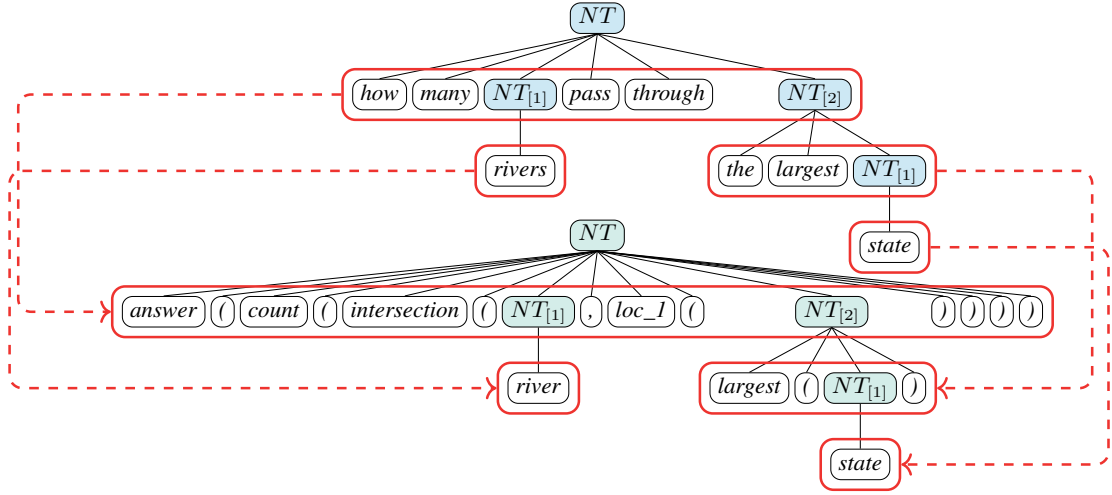
\begin{figure*}[t!]

\small{
$ NT \rightarrow \langle \textrm{how many}~NT_{\boxnum{1}}~\textrm{pass through}~NT_{\boxnum{2}}, \textrm{answer ( count ( intersection (}~NT_{\boxnum{1}}~\textrm{, loc\_1 (}~NT_{\boxnum{2}}~\textrm{) ) ) )} \rangle $ \\
$ NT \rightarrow \langle \textrm{rivers}, \textrm{river} \rangle $ \\ 
$ NT \rightarrow \langle \textrm{the largest}~NT_{\boxnum{1}}, \textrm{largest (}~NT_{\boxnum{1}}~\textrm{)} \rangle $ \\
$ NT \rightarrow \langle \textrm{state}, \textrm{state} \rangle $
}

\begin{center}
\scalebox{1.2}{

\begin{tikzpicture}[every node/.style={draw=black, rectangle,rounded corners, inner xsep=3pt, inner ysep=1pt},
snt/.style={fill=lblue},
tnt/.style={fill=lgreen}, scale=0.8]

\Tree [. \node[snt]{\strut $NT$};
    \node(s1s){\emph{\strut how}};
    \node{\emph{\strut many}};
    [. \node(s2)[snt]{\strut $NT_{\boxnum{1}}$};
      \node(s2s){\emph{\strut rivers}};
    ]
    \node{\emph{\strut pass}};
    \node{\emph{\strut through}};
    [. \node(s1e)[snt]{\strut $NT_{\boxnum{2}}$};
      \node(s3s){\emph{\strut the}};
      \node{\emph{\strut largest}};
      [. \node(s3e)[snt]{\strut $NT_{\boxnum{1}}$};
        \node(s4s){\emph{\strut state}};
      ]
    ]
  ]
]
\begin{scope}[yshift=-3cm]

\Tree [. \node[tnt]{\strut $NT$};
    \node(t1s){\emph{\strut answer}};
    \node{\emph{\strut (}};
    \node{\emph{\strut count}};
    \node{\emph{\strut (}};
    \node{\emph{\strut intersection}};
    \node{\emph{\strut (}};
    [. \node[tnt]{\strut $NT_{\boxnum{1}}$};
      \node(t2s){\emph{\strut river}};
    ]
    \node{\emph{\strut ,}};
    \node{\emph{\strut loc\_1}};
    \node{\emph{\strut (}};
    [. \node[tnt]{\strut $NT_{\boxnum{2}}$};
      \node(t3s){\emph{\strut largest}};
      \node{\emph{\strut (}};
      [. \node[tnt]{\strut $NT_{\boxnum{1}}$};
        \node(t4s){\emph{\strut state}};
      ]
      \node(t3e){\emph{\strut )}};
    ]
    \node{\emph{\strut )}};
    \node{\emph{\strut )}};
    \node{\emph{\strut )}};
    \node(t1e){\emph{\strut )}};
  ]
]
\end{scope}
\node(s1)[draw=dred,thick,inner sep=3pt,fit=(s1s)(s1e)]{};
\node(t1)[draw=dred,thick, inner sep=3pt,fit=(t1s)(t1e)]{};

\node(s2)[draw=dred,thick,inner sep=3pt,fit=(s2s)]{};
\node(t2)[draw=dred,thick, inner sep=3pt,fit=(t2s)]{};

\node(s3)[draw=dred,thick,inner sep=3pt,fit=(s3s)(s3e)]{};
\node(t3)[draw=dred,thick, inner sep=3pt,fit=(t3s)(t3e)]{};

\node(s4)[draw=dred,thick,inner sep=3pt,fit=(s4s)]{};
\node(t4)[draw=dred,thick, inner sep=3pt,fit=(t4s)]{};

% Lines connecting source to target.
\draw [->,draw=dred,thick,dashed,rounded corners] (s1) -- ++(-8,0) |- (t1);
\draw [->,draw=dred,thick,dashed,rounded corners] (s2) -- ++(-7,0) |- (t2);
\draw [->,draw=dred,thick,dashed,rounded corners] (s3) -- ++(3.7,0) |- (t3);
\draw [->,draw=dred,thick,dashed,rounded corners] (s4) -- ++(3,0) |- (t4);
\end{tikzpicture}
}
\end{center}
\caption{An example QCFG derivation.
Each non-terminal in the source derivation (blue) corresponds to a non-terminal in the target derivation (green). The QCFG rules used in the derivation are shown above.}
\label{parse_tree}
\end{figure*}

Synchronous Context-Free Grammars (SCFGs) have been used to model the hierarchical mapping between pairs of strings in areas such as compiler theory~\cite{aho1972theory} and natural language processing.

Informally, SCFGs can be viewed as an extension of Context-Free Grammars (CFGs) that \emph{synchronously} generate strings in both a source and target language. We write SCFG rules as:
$$ S \rightarrow \langle \alpha,\beta \rangle $$
Where $S$ is a non-terminal symbol, and $\alpha$ and $\beta$ are strings of non-terminal and terminal symbols. An SCFG rule can be viewed as two CFG rules, $S \rightarrow \alpha$ and $S \rightarrow \beta$ with a pairing between the occurrences of non-terminal symbols in $\alpha$ and $\beta$. This pairing is indicated by assigning each non-terminal in $\alpha$ and $\beta$ an index $\in \mathbb{N}$. Non-terminals sharing the same index are called \emph{linked}. Following convention, we denote the index for a non-terminal using a boxed subscript, e.g. $NT_{\boxnum{1}}$. A complete SCFG derivation is a pair of parse trees, one for the source language and one for the target language. An example derivation is shown in Figure ~\ref{parse_tree}.

The $\Rightarrow^r$ operator refers to a \emph{derives} relation, such that $\langle \alpha^1, \beta^1 \rangle \Rightarrow^r \langle \alpha^2, \beta^2 \rangle$ states that the string pair $\langle \alpha^2, \beta^2 \rangle$ can be generated from $\langle \alpha^1, \beta^1 \rangle$ by applying the rule $r$.
We write $\Rightarrow$ to leave the rule unspecified, assuming the set of possible rules is clear from context. We write $\Rightarrow\Rightarrow$ to indicate a chain of 2 rule applications, omitting the intermediate string pair.
Finally, we write $\stackrel{*}{\Rightarrow}$ to denote the reflexive transitive closure of $\Rightarrow$. 

\paragraph{Quasi-Synchronous Context-Free Grammars (QCFGs)}
QCFGs generalize SCFGs in various ways, notably relaxing the restriction on a strict one-to-one alignment between source and target non-terminals~\cite{smith2006quasi}. 

\paragraph{Compositionality} Notably, grammar formalisms such SCFGs and QCFGs capture the formal notion of the principle of compositionality as a homomorphism between source and target structures~\cite{montague1970universal,janssen1997compositionality}.

\subsection{\qg Grammar Induction Details}
\label{sec:qcfq_induction}

Having defined the codelength scoring function that we use to compare grammars in section~\ref{sec:nqg-grammar-induction}, we describe our greedy search algorithm that finds a grammar that approximately minimizes this objective.\footnote{The induction objective contains hyperparameters representing the bitlength of terminal and non-terminal symbols. For all experiments we use $l_{N} = 1$. For \textsc{GeoQuery} and \textsc{Spider} we use $l_T = 8$, and use $l_T = 32$ for \textsc{SCAN}.}

\paragraph{Initialization}
We initialize $\mathcal{R}$ to be $\{ NT \rightarrow \langle \mathbf{x}, \mathbf{y} \rangle \mid \mathbf{x}, \mathbf{y} \in \mathcal{D} \} $. We also add identity rules for substrings that exactly match between source and target examples, e.g. $ NT \rightarrow \langle k, k \rangle $ where $k$
is a substring of both $\mathbf{x}$ and $\mathbf{y}$ for some $\mathbf{x},\mathbf{y} \in \mathcal{D}$.\footnote{These initialization rules are used for \textsc{GeoQuery} and \textsc{Spider}, but \textsc{SCAN} does not contain any exact token overlap between source and target languages.} 

\paragraph{Optimization Algorithm}
Our algorithm was designed with simplicity in mind, and therefore uses a simple greedy search process that could likely be significantly improved upon by future work. At a high level, our greedy algorithm iteratively identifies a rule to be added to $\mathcal{R}$ that decreases the codelength by enabling $\geq 1$ rules in $\mathcal{R}$ to be removed while maintaining the invariant that $\mathcal{G}$ allows for deriving all of the training examples, i.e. $\langle NT, NT \rangle \stackrel{*}{\Rightarrow} \langle \mathbf{x}, \mathbf{y} \rangle$ for every $\mathbf{x},\mathbf{y} \in \mathcal{D}$. The search completes when no rule that decreases $L(\mathcal{R})$ can be identified.

To describe the implementation, first let us define several operations over rules and sets of rules. We define the set of rules that can be derived from a given set of rules, $\mathcal{R}$:
$$ d(\mathcal{R}) = \{ NT \rightarrow \langle \alpha, \beta \rangle \mid \langle NT, NT \rangle \stackrel{*}{\Rightarrow} \langle \alpha, \beta \rangle \}
$$

We define an operation $\mathrm{SPLIT}$ that generates possible choices for \emph{splitting} a rule into 2 rules:
\begin{equation*}
\begin{gathered}
\mathrm{SPLIT}(NT \rightarrow \langle \alpha, \beta \rangle) = \{ g, h \mid \\
\langle NT, NT \rangle \Rightarrow^g   \Rightarrow^h  \langle \alpha, \beta \rangle ~ \lor \\
\langle NT, NT \rangle \Rightarrow^h   \Rightarrow^g  \langle \alpha, \beta \rangle \},
\end{gathered}
\end{equation*}
where $g$ and $h$ is a pair of new rules that would maintain the invariant that $\langle NT, NT \rangle \stackrel{*}{\Rightarrow}  \langle \mathbf{x}, \mathbf{y} \rangle$ for every $\mathbf{x}, \mathbf{y} \in \mathcal{D}$, even if the provided rule is eliminated.\footnote{We optionally allow $\mathrm{SPLIT}$ to introduce repeated target non-terminals when the target string has repeated substrings. Otherwise, we do not allow $\mathrm{SPLIT}$ to replace a repeated substring with a non-terminal, as this can lead to an ambiguous choice. We enable this option for \textsc{SCAN} and \textsc{Spider} but not for \textsc{GeoQuery}, as FunQL does not require such repetitions.}

$\mathrm{SPLIT}$ can be implemented by considering pairs of sub-strings in $\alpha$ and $\beta$ to replace with a new indexed non-terminal symbol. For example, the rule ``$NT \rightarrow \langle \textrm{largest state}, \textrm{largest ( state )} \rangle$'' can be split into the rules ``$NT \rightarrow \langle \textrm{largest}~NT_{\boxnum{1}}, \textrm{largest (}~NT_{\boxnum{1}}~\textrm{)} \rangle$'' and ``$NT \rightarrow \langle \textrm{state}, \textrm{state} \rangle$''. This step can require re-indexing of non-terminals.

During our greedy search, we only split rules when one of the two resulting rules can already be derived given $\mathcal{R}$. Therefore, we define a function $\mathrm{NEW}$ that returns a set of candidate rules to consider:
\begin{equation*}
\begin{gathered}
\mathrm{NEW}(\mathcal{R}) = \\
\{ g \mid g, h \in \mathrm{SPLIT}(f) \land f \in \mathcal{R} \land h \in d(\mathcal{R}) \}
\end{gathered}
\end{equation*}

Similarly, we can compute the set of rules that are made redundant and can be eliminated by introducing one these candidate rules, $f$:
\begin{equation*}
\begin{gathered}
\mathrm{ELIM}(\mathcal{R}, f) = \\
\{ h \mid f, g \in \mathrm{SPLIT}(h) \land g \in d(\mathcal{R}) \land h \in \mathcal{R} \}
\end{gathered}
\end{equation*}

We can then define the codelength reduction of adding a particular rule, $
-\Delta L(\mathcal{R}, f) = L(\mathcal{R}) - L(\mathcal{R'})$ where $\mathcal{R'} = (\mathcal{R} ~\cup~ f) \setminus \mathrm{ELIM}(\mathcal{R}, f)$.\footnote{The last term of the codelength objective described in section~\ref{sec:nqg-grammar-induction} is related to the increase in the proportion of incorrect derivations due to introducing $f$. Rather than computing this exactly, we estimate this quantity by sampling up to $k$ examples from $\mathcal{D}$ that contain all of the sub-strings of source terminal symbols in $f$ such that $f$ could be used in a derivation, and estimating the increase in incorrect derivations over this sample only. We sample $k=10$ examples for all experiments.}
Finally, we can select the rule with the largest $-\Delta L$:
$$ \mathrm{MAX}(\mathcal{R}) =  \underset{f \in \mathrm{NEW}(\mathcal{R})}{\mbox{argmax}} -\Delta L (\mathcal{R}, f) $$

Conceptually, after initialization, the algorithm then proceeds as:

\begin{algorithm}
\begin{algorithmic}
\WHILE{ $ |\mathrm{NEW}( \mathcal{R} )| > 0 $ }
    \STATE $ r \gets \mathrm{MAX}(\mathcal{R}) $
    \IF { $ -\Delta L (\mathcal{R}, r) < 0$ }
    \STATE \bf{break}
    \ENDIF
    \STATE $ \mathcal{R} \gets (\mathcal{R} ~\cup~ r) \setminus \mathrm{ELIM}(\mathcal{R}, r) $
\ENDWHILE
\end{algorithmic}
\end{algorithm}

For efficiency, we select the shortest $N$ examples from the training dataset, and only consider these during the induction procedure. Avoiding longer examples is helpful as the number of candidates returned by $\mathrm{SPLIT}$ is polynomial with respect to source and target length.  Once induction has completed, we then determine which of the longer examples cannot be derived based on the set of induced rules, and add rules for these examples.\footnote{We use $N=500$ for \textsc{SCAN} and $N=1000$ for \textsc{Spider}. As the \textsc{GeoQuery} training set contains $<500$ unique examples, we use the entire training set.}

Our algorithm maintains a significant amount of state between iterations to cache computations that are not affected by particular rule changes, based on overlap in terminal symbols. We developed the algorithm and selected some hyperparameters by assessing the size of the induced grammars over the training sets of \textsc{SCAN} and \textsc{GeoQuery}.

Our grammar induction algorithm is similar to the transduction grammar induction method for machine translation by \citet{saers-etal-2013-unsupervised}. More broadly, compression-based criteria have been successfully used by a variety of models for language \cite{grunwald1995minimum,tang2001using,ravi-knight-2009-minimized,poon-etal-2009-unsupervised}. 

\subsection{\qg Parsing Model Details}
\label{sec:appendix-parsing-model}

In this section we provide details on how we generate derivation scores, $s(\mathbf{z}, \mathbf{x})$, using a neural model, as introduced in \S~\ref{sec:nqg-component}. The derivation scores decompose over anchored rules from our grammar:
$$ s(\mathbf{z}, \mathbf{x}) = \sum\limits_{(r,i,j) \in \mathbf{z}} \phi(r,i,j,\mathbf{x}), $$
where $r$ is an index for a rule in $\mathcal{G}$ and $i$ and $j$ are indices defining the anchoring in $\mathbf{x}$. The anchored rule scores, $\phi(r,i,j,\mathbf{x})$, are based on contextualized representations from a BERT~\cite{devlin2018bert} encoder:
$$ \phi(r,i,j,\mathbf{x}) = f_s([w_i, w_j]) + e_r^\intercal f_r([w_i, w_j]), $$
where $[w_i, w_j]$ is the concatenation of the BERT representations for the first and last wordpiece in the anchored span, $f_r$ is a feed-forward network with hidden size $d$ that outputs a vector $\in \mathbb{R}^d$, $f_s$ is a feed-forward network with hidden size $d$ that outputs a scalar, and $e_r$ is an embedding $\in \mathbb{R}^d$ for the rule index $r$. Our formulation for encoding spans is similar to that used in other neural span-factored models \cite{stern2017Minimal,lee2017end}.

\section{Experimental Details}
\label{sec:appendix-exp-details}

\subsection{Model Hyperparameters and Runtime}

We selected reasonable hyperparameter values and performed some minimal hyperparameter tuning for T5 and \qg based on random splits of the training sets for {\textsc{GeoQuery}} and {\textsc{Spider}}. We used the same hyperparameters for all splits of a given dataset.

For T5, we selected a learning rate of $1e^{-4}$ from $[1e^{-3}, 1e^{-4}, 1e^{-5}]$, which we used for all experiments. Otherwise, we used the default hyperparameters for fine-tuning. We fine-tune for $3,000$ steps for {\textsc{GeoQuery}} and $10,000$ for {\textsc{Spider}}. T5-Base trained with a learning rate of $1e^{-4}$ reached 94.2\% accuracy at $3,000$ steps on a random split of the standard GeoQuery training set into 500 training and 100 validation examples.

\begin{table*}[!t]
\begin{center}

\scalebox{0.85}{

\begin{tabular}{p{1.0\linewidth}}
\toprule

\emph{\bf Source:} how many states are next to major rivers \\
\emph{\bf Target:} \nlp{answer ( count ( intersection ( state , next\_to\_2 ( intersection ( major , river ) ) ) ) )} \\
\emph{\bf Prediction:} \nlp{answer ( count ( intersection ( state , next\_to\_2 ( intersection ( major , \textcolor{red}{intersection ( river , m0 )} ) ) ) ) )} \\
\emph{\bf Notes:} The trigram ``\nlp{major , intersection}'' occurs 28 times during training, but ``\nlp{major , river}'' occurs 0 times. In this case, T5 also hallucinates ``\nlp{m0}'' despite no entity placeholder occuring the source. \\

\midrule

\emph{\bf Source:} which state has the highest peak in the country \\
\emph{\bf Target:} \nlp{answer ( intersection ( state , loc\_1 ( highest ( place ) ) ) ) }\\
\emph{\bf Prediction:} \nlp{answer ( \textcolor{red}{highest (} intersection ( state , \textcolor{red}{loc\_2} ( highest ( \textcolor{red}{intersection ( mountain , loc\_2 ( m0 ) ) )} ) ) )}\\
\emph{\bf Notes:} The token ``\nlp{highest}'' occurs after ``\nlp{answer (}'' in 83\% of instances in which ``\nlp{highest}'' occurs in the training set. Note that T5 also hallucinates ``\nlp{m0}'' in this case. \\
\bottomrule
\end{tabular}
}

\caption{Example prediction errors for T5-Base for the \textsc{GeoQuery} \tmcd split.}
\label{tab:geoquery_errors}

\end{center}
\end{table*}
\begin{table}[t]
\begin{center}

\scalebox{0.75}{
\begin{tabular}{lccc}
\toprule
Dataset & Examples & Induced Rules & Ratio  \\
\midrule
{\textsc{SCAN}} & 16727 & 21 & 796.5 \\
{\textsc{GeoQuery}} & 600 & 234 & 2.6 \\
{\textsc{Spider-SSP}} & 3282 & 4155 & 0.79 \\

\bottomrule
\end{tabular}
}

\caption{Sizes of induced grammars.}
\label{grammar_sizes}

\end{center}
\end{table}

\begin{table}[t]
\begin{center}

\scalebox{0.75}{

\begin{tabular}{lcccc}
\toprule
& Std. & Templ. & Len. & \tmcd \\
\midrule
\qgxt-3B Acc. & 0.6 & 1.2 & 1.2 & 0.4 \\
\qgxt-Base Acc. & 0.5 & 1.4 & 1.1 & 0.4 \\
\qg Acc. & 1.2 & 4.5 & 1.5 & 0.4 \\
\midrule
\qg Coverage & 0.7 & 3.4 & 1.8 & 0.1 \\
\qg Precision & 0.7 & 1.9 & 1.7 & 1.2 \\
\bottomrule
\end{tabular}
}

\caption{Standard deviation of \qg for \textsc{GeoQuery}.}
\label{tab:geoquery_variance}

\end{center}
\end{table}

\eat{
\begin{table*}[!t]
\begin{center}

\scalebox{0.82}{

\begin{tabular}{llccccc}
\toprule
Split & & \qg Precision & \qg Recall & \qgxt-3B Acc. & \qgxt-Base Acc. & \qg Acc. \\
\midrule
Standard	&	Run 1	&	95.5	&	79.6	&	92.5	&	93.2	&	76.1	\\
	&	Run 2	&	95.1	&	80.0	&	92.5	&	93.6	&	76.1	\\
	&	Run 3	&	96.5	&	81.1	&	93.6	&	94.3	&	78.2	\\
	&	Mean	&	95.7	&	80.2	&	92.9	&	93.7	&	76.8	\\
	&	Std. Dev.	&	0.7	&	0.7	&	0.6	&	0.5	&	1.2	\\
\midrule
Length	&	Run 1	&	86.3	&	41.4	&	50.9	&	50.2	&	35.7	\\
	&	Run 2	&	84.8	&	44.8	&	52.5	&	51.4	&	38.0	\\
	&	Run 3	&	88.1	&	43.9	&	53.2	&	52.5	&	38.6	\\
	&	Mean	&	86.4	&	43.3	&	52.2	&	51.4	&	37.4	\\
	&	Std. Dev.	&	1.7	&	1.8	&	1.2	&	1.1	&	1.5	\\
\midrule
MCD	&	Run 1	&	94.8	&	43.6	&	56.8	&	54.3	&	41.4	\\
	&	Run 2	&	92.7	&	43.9	&	56.1	&	53.6	&	40.7	\\
	&	Run 3	&	94.8	&	43.6	&	56.8	&	54.3	&	41.4	\\
	&	Mean	&	94.1	&	43.7	&	56.6	&	54.1	&	41.1	\\
	&	Std. Dev.	&	1.2	&	0.1	&	0.4	&	0.4	&	0.4	\\
\midrule
Template	&	Run 1	&	96.9	&	66.3	&	89.5	&	85.9	&	64.2	\\
	&	Run 2	&	93.6	&	60.6	&	87.5	&	83.4	&	56.7	\\
	&	Run 3	&	96.9	&	66.7	&	89.5	&	85.9	&	64.7	\\
	&	Mean	&	95.8	&	64.5	&	88.8	&	85.0	&	61.9	\\
	&	Std. Dev.	&	1.9	&	3.4	&	1.2	&	1.4	&	4.5	\\
\bottomrule
\end{tabular}
}

\caption{Results from different runs of training \qg for \textsc{GeoQuery}.}
\label{tab:geoquery_variance}

\end{center}
\end{table*}

}

For the \qg neural model, we use the pre-trained BERT Tiny model of \citet{turc2019well} (4.4M parameters) for \textsc{SCAN} and \textsc{Spider}, and BERT Base (110.1M parameters) for \textsc{GeoQuery}, where there is more headroom for improved scoring. We do not freeze pre-trained BERT parameters during training. For all experiments, we use $d=256$ dimensions for computing anchored rule scores. We fine-tune for $256$ steps and use a learning rate of $1e^{-4}$. We use a batch size of $256$. 

We train \qg on 8 V100 GPUs. Training \qg takes < 5 minutes for \textsc{SCAN} and \textsc{Spider} (BERT Tiny), and up to 90 minutes for \textsc{GeoQuery} (BERT Base). We fine-tune T5 on 32 Cloud TPU v3 cores.\footnote{https://cloud.google.com/tpu/} For \textsc{GeoQuery}, fine-tuning T5 takes approximately 5 and 37 hours for Base and 3B, respectively. For \textsc{Spider}, fine-tuning T5 takes approximately 5 and 77 hours for Base and 3B, respectively.

\subsection{Dataset Preprocessing}

For \textsc{GeoQuery}, we use the version of the dataset with variable-free FunQL logical forms~\cite{kate2005learning}, and expand certain functions based on their logical definitions, such that $\nlp{state(next\_to\_1(state(all)))}$ becomes the more conventional $\nlp{intersection(state, next\_to\_1(state))}$. We replace entity mentions with placeholders (e.g. ``\nlp{m0}'', ``\nlp{m1}'') in both the source and target. 

For \textsc{Spider}, we prepend the name of the target database to the source sequence. For T5, we also serialize the database schema as a string and append it to the source sequence similarly to~\citet{suhr2020exploring}. This schema string contains the names of all tables in the database, and the names of the columns for each table. As we use a maximum source sequence length of 512 for T5, this leads to some schema strings being truncated (affecting about 5\% of training examples).

\textsc{SCAN} did not require any dataset-specific preprocessing. 

\subsection{Atom and Compound Definitions}
\label{sec:appendix-tmcd-definitions}

For \textsc{GeoQuery}, the tree structure of FunQL is given by explicit bracketing. We define atoms as individual FunQL symbols, and compounds as combinations between parent and child symbols in the FunQL tree. Example atoms are 
$\nlp{longest}$, $\nlp{river}$, and  $\nlp{exclude}$ and example compounds are $\nlp{longest(river)}$ and $\nlp{exclude(longest(\_), \_)}$.

For \textsc{Spider}, we tokenize the SQL string and define atoms as individual tokens.
To define compounds, we parse the SQL string using an unambiguous CFG, and define compounds from the resulting parse tree. We define compounds over both first and second order edges in the resulting parse tree.

\subsection{Grammar Sizes}
\label{sec:appendix-grammar-sizes}

Induced grammar sizes for a selected split of each dataset are shown in Table~\ref{grammar_sizes}. 
For ~\textsc{Spider}, the number of induced rules is larger than the original dataset due to the identity rules added during initialization. 

\subsection{\textsc{GeoQuery} Variance}
\label{sec:appendix-geoquery-variance}

In tables \ref{tab:scan_geo} and \ref{tab:qg_precision} we report the mean of 3 runs for \qg for \textsc{GeoQuery}. The standard deviations for these runs are reported in Table \ref{tab:geoquery_variance}. The reported standard deviations for \qgxt use the same fine-tuned T5 checkpoint, so they do not reflect any additional variance from different fine-tuned T5 checkpoints.

\subsection{T5 \textsc{GeoQuery} Errors}
\label{sec:appendix-geoquery-examples}

We include several example T5-Base errors on the \textsc{GeoQuery} \tmcd split in Table~\ref{tab:geoquery_errors}.

\end{document}